%% file: arxiv.tex
\documentclass{article}
\usepackage{graphicx} 
\usepackage{a4wide}
\usepackage{amsmath,amsfonts,amssymb,amsthm}
\usepackage{nicefrac}
\usepackage{cite}
\usepackage{nicefrac}
\usepackage{authblk}

\usepackage{algorithm}
\usepackage[noend]{algcompatible}
\algnewcommand\algorithmicreturn{\textbf{Return}}
\algnewcommand\RETURN{\State \algorithmicreturn}%

\usepackage{graphicx}
\usepackage{textcomp}
\usepackage{xcolor}

\def\BibTeX{{\rm B\kern-.05em{\sc i\kern-.025em b}\kern-.08em
    T\kern-.1667em\lower.7ex\hbox{E}\kern-.125emX}}

\input{macros.tex}

\newtheorem{assumption}{Assumption}

\newtheorem{definition}{Definition}
\newtheorem{theorem}{Theorem}

\title{Distributed Gradient Clustering: Convergence and the Effect of Initialization}

\author[1]{Aleksandar Armacki\thanks{The first two authors contributed equally.}\thanks{The work of A. Armacki and S. Kar is supported in part by the National Science Foundation under grant ECCS 2330196. The work of H. Sharma was supported in part by a SPARC grant from the MHRD, Govt. of India. The work of D. Bajovi\'{c} is supported by the Ministry of Science, Technological Development and Innovation (Contract No. 451-03-65/2024-03/200156), by Faculty of Technical Sciences, University of Novi Sad, through project ``Scientific and Artistic Research Work of Researchers in Teaching and Associate Positions at the Faculty of Technical Sciences, University of Novi Sad'' (No. 01-3394/1), and by Serbian Ministry of Science, Technological development and Innovation, within the bilateral project Serbia-Slovakia No. 337-00-3/2024-05/16. The work of D. Jakoveti\'{c} is supported by the Science Fund of Republic of Serbia, project ``LASCADO'', grant No. 7359, by Provincial Secretariat for Higher Education and Scientific Research, grant No. 142-451- 2593/2021-01/2. The work of D. Jakoveti\'{c} and D. Bajovi\'{c} is also supported by the European Union’s Horizon Europe program under grant agreement No. 101093006.  
 \emph{(Corresponding author: Aleksandar Armacki.)}}
}
\author[2]{Himkant Sharma$^*$}
\author[3]{Dragana Bajovi\'{c}}
\author[4]{Du\v{s}an Jakoveti\'{c}}
\author[2]{Mrityunjoy Chakraborty}
\author[1]{Soummya Kar}
\affil[1]{Carnegie Mellon University, Pittsburgh, Pennsylvania, USA \newline \texttt{\{aarmacki,soummyak\}@andrew.cmu.edu}}
\affil[2]{Indian Institute of Technology Kharagpur, Kharagpur, India \newline \texttt{himkant@kgpian.iitkgp.ac.in, mrityun@ece.iitkgp.ac.in}}
\affil[3]{Faculty of Technical Sciences, University of Novi Sad, Novi Sad, Serbia \newline \texttt{dbajovic@uns.ac.rs}}
\affil[4]{Faculty of Sciences, University of Novi Sad, Novi Sad, Serbia \newline \texttt{dusan.jakovetic@dmi.uns.ac.rs}}

\date{}
    
\begin{document}

\maketitle

\begin{abstract}
    We study the effects of center initialization on the performance of a family of distributed gradient-based clustering algorithms introduced in \cite{armacki2024unified}, that work over connected networks of users. In the considered scenario, each user contains a local dataset and communicates only with its immediate neighbours, with the aim of finding a global clustering of the joint data. We perform extensive numerical experiments, evaluating the effects of center initialization on the performance of our family of methods, demonstrating that our methods are more resilient to the effects of initialization, compared to centralized gradient clustering \cite{pmlr-v162-armacki22a}. Next, inspired by the $K$-means++ initialization \cite{kmeans++}, we propose a novel distributed center initialization scheme, which is shown to improve the performance of our methods, compared to the baseline random initialization.
\end{abstract}

\section{Introduction}

Clustering is an unsupervized learning problem, with the goal of finding groups of similar data, without having any knowledge of the underlying distribution, or even the true number of groups \cite{clustering-survey,JAIN2010651}. Depending on the approach, data can be assigned to clusters in a \emph{hard} or \emph{soft} manner, with hard clustering assigning data to exclusively one cluster, while soft clustering provides the probability of a sample belonging to each cluster. In this paper we will be studying the problem of hard clustering. Additionally, we will focus on center-based clustering, e.g., \cite{awasthi2016center}, where the goal is to find centers which represent the clusters. Many popular algorithms fall in this category, including the celebrated Lloyd's method \cite{Lloyd}, its extension to Bregman losses \cite{JMLR:v6:banerjee05b}, Huber loss clustering \cite{Huber_clust}, as well as the recently proposed gradient based clustering \cite{pmlr-v162-armacki22a}. 

Clustering has traditionally been studied in the centralized regime, where the methods are run across the entire dataset. Another learning paradigm, which has been attracting significant interest is that of \emph{distributed learning}. \emph{Distributed learning} is a popular learning paradigm, wherein many users collaborate to train a joint model, while keeping their data private. There are many approaches to distributed learning, such as federated learning (FL), e.g., \cite{pmlr-v54-mcmahan17a,kairouz-fl,armacki-oneshot,armacki-personalized}, and peer-to-peer (P2P) distributed learning, e.g., \cite{vlaski_et_al,sayed-networks,jakovetic-fast,nedic-subgrad}. In this paper we are interested in the P2P setup, where users communicate directly with one another, while no user can communicate directly with all the others. The communication network is modeled as a connected graph $G = (V,E)$. Clustering in this setup is very challenging, as the data is stored locally at each user, with users only able (or willing) to exchange local parameter estimates (e.g., local centers) in order to achieve the final goal of obtaining a clustering of the entire, joint dataset.

\textbf{Literature review.} Distributed clustering has been considered in \cite{kmean-dynamic,dist-clust-wsn,dist-coresets,oliva2013distributed,dist-fuzzy,kar2019clustering,dist-soft}. Work \cite{kmean-dynamic} proposes approximate $K$-means algorithms for both P2P and FL setups, providing theoretical guarantees only in the FL setup. Works \cite{dist-clust-wsn,dist-fuzzy,dist-soft} study distributed soft and hard $K$-means clustering, with only the method in \cite{dist-clust-wsn} providing convergence guarantees to a local minima of the centralized $K$-means problem. In \cite{dist-coresets}, the authors study $K$-means and $K$-medians problems and rely on the idea of coresets \cite{coresets}, to design methods with provable constant approximation guarantees. Work \cite{oliva2013distributed} studies distributed $K$-means in the special case where users have a single sample, while in \cite{kar2019clustering} the authors design a parametric family of distributed $K$-means methods, establishing convergence of centers to local minima of the centralized $K$-means problem. Finally, we propose a unified framework for distributed clustering in \cite{armacki2024unified}, that considers popular clustering methods beyond $K$-means, such as Huber loss clustering \cite{Huber_clust}. 

\textbf{Contributions.} In this work we study the effects of center initialization on the performance of the distributed gradient-based clustering (DGC-$\mathcal{F}_\rho$) method proposed in \cite{armacki2024unified}. To that end, we perform extensive numerical experiments, demonstrating that DGC-$\mathcal{F}_\rho$ is more resilient to center initialization, compared to the centralized gradient clustering (CGC) method from \cite{pmlr-v162-armacki22a}. Inspired by the celebrated $K$-means++ initialization, we then propose a novel distributed center initialization scheme, dubbed Distributed $K$-means+Clustering (DKM+C), which combines local $K$-means++ with multiple communication and local clustering rounds, to produce the initial centers. The proposed scheme is shown to result in better performance of the algorithm, compared to the baseline random center initialization.    

\textbf{Paper organization.} The rest of the paper is organized as follows. Section \ref{sec:problem} formally states the problem of distributed center-based clustering, Section \ref{sec:method} introduces the proposed family of methods, Section \ref{sec:theory} provides theoretical results, Section \ref{sec:num} provides numerical results and Section \ref{sec:conclusion} concludes the paper. The remainder of this section introduces notation.

\textbf{Notation.} The spaces of real numbers and $d$-dimensional vectors are denoted by $\R$ and $\R^d$, with $\|\cdot\|$ denoting the Euclidean norm. The set of non-negative integers is denoted by $\N$, with $[M] = \{1,\ldots,M\}$, for any $M \in \N$. For a matrix $A \in \R^{d \times d}$, $A^\top$ and $\overline{\lambda}(A)$ denote transposition and the largest eigenvalue of $A$. Superscripts and subscripts denote iterations and users, while brackets correspond to the particular center or cluster, e.g., $x^t_i(k)$ is center $k$ of user $i$ at iteration $t$.

\section{Problem Formulation}\label{sec:problem}

Consider a network of $m > 1$ users, communicating over a graph $G = (V,E)$, where $V = [m]$ is the set of vertices (i.e., users), $E$ is the set of undirected edges connecting them, such that $\{i,j\} \in E$ if and only if users $i$, $j$ communicate. Each user contains a local dataset $\D_i = \{y_{i,1},\ldots,y_{i,N_i} \} \subset \R^d$, for some $N_i \geq 1$. The goal is to produce a clustering of the global data $\D = \cup_{i \in [m]}\D_i$, into $K \geq 2$ disjoint clusters. Formally, the problem can be stated as
\begin{equation}\label{eq:general-constr}
    \min_{\substack{\bx_i \in \R^{Kd}, \: C_i \in \C_{K,\D_i}, \: i \in [m] \\ \text{subject to }\bx_1 = \ldots = \bx_m}}\sum_{i \in [m]}\sum_{k \in [K]}\sum_{r \in C_i(k)}\hspace{-0.5em}f(x_i(k),y_{i,r}),
\end{equation} where $\bx_i = \begin{bmatrix}x_i(1)^\top & \ldots & x_i(K)^\top  \end{bmatrix}^\top$ is the vector stacking the $K$ centers $x_i(k) \in \R^d$ of user $i$, $\C_{K,\D_i}$ is the set of all $K$-partitions of $\D_i$, i.e., $C_i \in \C_{K,\D_i}$ is a $K$-tuple $C_i = \begin{pmatrix}C_i(1), \ldots, C_i(K) \end{pmatrix}$, such that $C_i(k) \subseteq \D_i$,\footnote{In a slight abuse of notation, we will also use $\D_i$ to denote the set of indices of the data, i.e., $\D_i = [N_i]$.} $C_i(k) \cap C_i(l) = \emptyset$ and $\cup_{k \in [K]}C_i(k) = \D_i$. Here $f: \R^d \times \R^d \mapsto [0,\infty)$ is a loss function, e.g., $f(x,y) = \|x - y\|^2$ recovers the distributed $K$-means problem, with many other possibilities, such as Huber, Logistic or Fair loss, see \cite{armacki2024unified}. In general, \eqref{eq:general-constr} is NP-hard, even in the centralized setting, e.g., \cite{kmeans-stability,Vattani2010TheHO,awasthi2015hardness}. As such, the best one can hope for is reaching stationary points, with various schemes guaranteeing this in both centralized, e.g., \cite{Macqueen67somemethods,Lloyd,JMLR:v6:banerjee05b,pmlr-v162-armacki22a}, and distributed settings, e.g., \cite{dist-clust-wsn,kar2019clustering,armacki2024unified}. 

The problem \eqref{eq:general-constr} ensures clustering of the joint data is produced, by requiring centers across all users to be the same. As \eqref{eq:general-constr} is a constrained problem, a relaxation is proposed in \cite{armacki2024unified}, making it amenable to a distributed first-order approach. The relaxed problem is given by
\begin{align}
    \min_{\substack{\bx \in \R^{Kmd}, \\ C \in \C_{m,K,\D}}} \hspace{-0.5em}J_\rho(\bx,C) &= \sum_{i \in [m]}\sum_{k \in [K]}\Big[ \frac{1}{2}\sum_{j \in \calN_i}\|x_i(k) - x_j(k)\|^2 \nonumber \\ &+ \frac{1}{\rho}\sum_{r \in C_i(k)}w_{i,r}f(x_i(k),y_{i,r}) \Big], \label{eq:general-decentr}
\end{align} where $\mathcal{C}_{m,K,\D}$ is the set of all clusterings of the entire data, i.e., for $C \in \mathcal{C}_{m,K,\D}$, we have $C = (C_1,\ldots,C_m)$, with $C_i \in \mathcal{C}_{K,\D_i}$, $\calN_i = \left\{j \in V: \{i,j \} \in E\right\}$ is the set of neighbours of user $i$ (not including $i$), while $\rho \geq 1$ is a tunable parameter. The formulation \eqref{eq:general-decentr} relaxes \eqref{eq:general-constr}, by considering an unconstrained problem which penalizes the difference of centers among neighbouring users and controls the trade-off between center estimation and proximity, via the parameter $\rho$. 

\section{The DGC-$\mathcal{F}_\rho$ Family of Methods}\label{sec:method}

In this section we describe the DGC-$\mathcal{F}_\rho$ family of methods proposed in \cite{armacki2024unified}. We refer to DGC-$\mathcal{F}_\rho$ as a family of methods, as it subsumes several distributed clustering methods, such as $K$-means, Huber, Logistic and Fair loss-based. In each iteration users maintain their center and cluster estimates. To begin, users choose initial centers $\bx_i^0 \in \R^{Kd}$, $i \in [m]$. At iteration $t \geq 0$, users first form the clusters locally, by finding a $k \in [K]$ for each data point $r \in \D_i$, such that the $k$-th center is the closest to the point $r$, i.e., such that 
\begin{equation}\label{eq:reassign}
    \|x_i^t(k) - y_{i,r}\| \leq \| x_i^t(l) - y_{i,r}\|, \: \text{for all } l \ne k,
\end{equation} and assign $y_{i,r}$ to $C_i^{t+1}(k)$. Next, the centers are updated via
\begin{align}
    x_i^{t+1}(k) = x_i^{t}(k) &- \alpha\sum_{j \in \calN_i} \left[x^{t}_i(k) - x^{t}_j(k) \right] \nonumber \\ &- \frac{\alpha}{\rho}\sum_{r \in C_i^{t+1}(k)} \nabla_x f\left(x_i^{t}(k),y_{i,r}\right), \label{eq:grad_local}
\end{align}
where $\alpha > 0$ is a fixed step-size. The procedure is summarized in Algorithm \ref{alg:dist-grad-cl}.\footnote{The method in \cite{armacki2024unified} is more general, in that it allows for distance metrics beyond Euclidean and multiple center updates per iteration, see \cite{armacki2024unified} for details.} Note that centers can be initialized randomly, providing flexibility in designing initialization algorithms, such as distributed variants of $K$-means++, e.g., \cite{dist-fuzzy}, or the method we propose in Section \ref{sec:num} ahead. The center update is built on the consensus+innovation framework, e.g., \cite{consensus+innovation1,consensus+innovation2}. 

\begin{algorithm}[!tb]
\caption{DGC-$\mathcal{F}_\rho$}
\label{alg:dist-grad-cl}
\begin{algorithmic}[1]
   \REQUIRE{$\alpha > 0$, $\rho \geq 1$, initial centers $\bx_i^0 \in \R^{Kd}$, $i \in [m]$.}
   \FOR{all users $i$ in parallel, in round t = 0,1,\ldots,T-1}
        \STATE {Set $C_i^{t+1}(k) \leftarrow \emptyset$, for all $k \in [K]$}; 
        \FOR{each $r \in [N_i]$}
            \STATE Find $k$ so that $\|x_i^t(k)-y_{i,r}\| \leq \|x_i^t(l)-y_{i,r}\|$, $l \neq k$;
            \STATE Update $C_i^{t+1}(k) \leftarrow C_i^{t+1}(k) \cup \{r\}$;
        \ENDFOR
        \STATE Exchange centers with neighbours $j \in \calN_i$;
        \STATE Update $x_i^{t+1}(k)$ by performing \eqref{eq:grad_local}, for all $k \in [K]$;
   \ENDFOR
   \RETURN{} $(\bx_i^T,C_i^T)$, $i \in [m]$.
    \end{algorithmic}
\end{algorithm}

\section{Convergence Guarantees}\label{sec:theory}

We start by defining the notion of points to which DGC-$\mathcal{F}_\rho$ converges to, referred to as \emph{fixed points}.

\begin{definition}\label{def:U}
Let $\bx \in \R^{Kmd}$ be cluster centers. We say that $U_\bx \subset \mathcal{C}_{m,K,\D}$ is the set of optimal clusterings with respect to $\bx$, if condition \eqref{eq:reassign} is satisfied for all clusterings $C \in U_\bx$.
\end{definition}

\begin{definition}\label{def:fix-pt}
The pair $(\bx^\star,C^\star) \in \R^{Kmd} \times \mathcal{C}_{m,K,\D}$ is a fixed point of DGC-$\mathcal{F}_\rho$, if 1) $C^\star \in U_{\bx^\star}$; 2) $\nabla J_\rho(\bx^\star,C^\star) = 0$.
\end{definition}

\begin{definition}\label{def:Ubar}
$\overline{U}_\bx \subset \mathcal{C}_{m,K,\D}$ is the set of clusterings, such that 1) $\overline{U}_\bx \subseteq U_\bx$; 2) $\nabla J_\rho(\bx,C) = 0$, for all $ C \in \overline{U}_\bx$.   
\end{definition}

Note that Definition \ref{def:fix-pt} requires $(\bx^\star,C^\star)$ to be a stationary point of $J_\rho$, in the sense that clusters $C^\star$ are optimal for fixed centers $\bx^\star$ and centers $\bx^\star$ are optimal for fixed clusters $C^\star$. As such, it is not possible to further improve the clusters, nor the centers at a fixed point. By Definitions \ref{def:U}-\ref{def:Ubar}, a point $\bx$ is a fixed point if and only if $\overline{U}_{\bx} \neq \emptyset$. As such, we will call a point $\bx$ a fixed point if $\overline{U}_{\bx} \neq \emptyset$. We next state our assumptions.

\begin{assumption}\label{asmpt:data}
    The full data has at least $K$ distinct samples.
\end{assumption}

\begin{assumption}\label{asmpt:graph}
    The graph $G = (V,E)$ is connected.
\end{assumption}

\begin{assumption}\label{asmpt:coerc} 
    The loss $f$ is coercive, convex and $\beta$-smooth with respect to the first argument and preserves the ordering with respect to Euclidean distance, i.e., for each $x,y,z \in \R^d$ 1) $\lim_{\|x\| \rightarrow \infty}f(x,y) = \infty$; 2) $0 \leq f(x,y) - f(z,y) - \langle \nabla_z f(z,y), x - z\rangle \leq \frac{\beta}{2}\|x - z\|^2$; 3) $f(x,y) < f(z,y)$ if $\|x-y\| < \|z-y\|$ and $f(x,y) = f(z,y)$ if $\|x-y\| = \|z-y\|$.
\end{assumption}

Assumptions \ref{asmpt:data}-\ref{asmpt:coerc} are mild assumptions on the global data, communication graph and loss function. Assumption \ref{asmpt:data} requires the full data to have $K$ distinct points, while placing no requirements on the local datasets. Assumption \ref{asmpt:graph} requires the communication graph to be connected, which allows for \eqref{eq:general-constr} to be solved in a distributed manner. Finally, Assumption \ref{asmpt:coerc} is a mild assumption on the behaviour of the loss, which is an intrinsic property of the loss, independent of the data that we wish to cluster. It is shown to be satisfied by a broad class of popular clustering losses, including $K$-means (i.e., squared Euclidean), Huber, Logsitic and Fair loss, see \cite{armacki2024unified} for details. We are now ready to state the convergence result from \cite{armacki2024unified}. 

\begin{theorem}\label{thm:convergence}
    Let Assumptions \ref{asmpt:data}-\ref{asmpt:coerc} hold. For the step-size $\alpha < (\nicefrac{\beta}{\rho} + \overline{\lambda}(L))^{-1}$, any initialization $\bx^0 \in \R^{Kmd}$ and $\rho \geq 1$, the sequence of centers $\{\bx^t\}_{t \in \N}$ generated by DGC-$\mathcal{F}_\rho$ converges to a fixed point $\bx^\star = \bx^\star(0,\rho) \in \R^{Kmd}$, such that $\overline{U}_{\bx^\star} \neq \emptyset$. Moreover, the clusters converge in finite time, i.e., there exists a $t_0 > 0$ such that $U_{\bx^t} \subseteq U_{\bx^\star}$, for all $t \geq t_0$. 
\end{theorem}

Theorem \ref{thm:convergence} shows that the sequence of centers generated by DGC-$\mathcal{F}_\rho$ is guaranteed to converge to a fixed point, for any center initialization. We emphasize that the fixed point to which the sequence of centers converges, $\bx^\star = \bx^\star(0,\rho)$, depends on center initialization and penalty parameter $\rho$. In the next section we perform extensive empirical studies of the effect of center initialization on the performance of DGC-$\mathcal{F}_\rho$. A detailed theoretical study on the effect of parameter $\rho$ on fixed points of DGC-$\mathcal{F}_\rho$, as $\rho \rightarrow \infty$, is provided in \cite{armacki2024unified}.

\section{Numerical Results}\label{sec:num}

\begin{figure}[tp]
    \centering
    \begin{tabular}{cc}
        \includegraphics[width=0.45\linewidth]{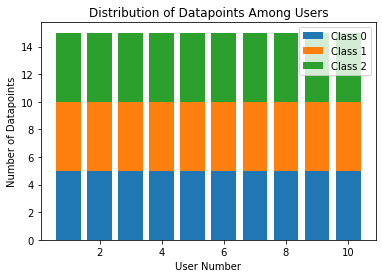}
        &
        \includegraphics[width=0.45\linewidth]{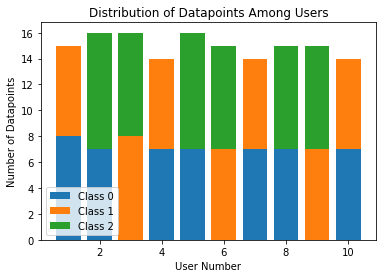}
    \end{tabular}
    \caption{Homogeneous and heterogeneous data distributions across users.}
    \label{fig:data}
\end{figure}

In this section we study the effect of center initialization of the performance of DGC-$\mathcal{F}_\rho$. In particular, we test the performance of our method using $K$-means loss, dubbed DGC-KM. All experiments are performed on Iris data \cite{iris}, which consists of $150$ samples, belonging to $K = 3$ classes ($50$ samples per class), and dimension $d = 4$. We distribute the samples across a ring network of $m = 10$ users. We consider two types of data distributions across users: \emph{homogeneous} and \emph{heterogeneous}. In the homogeneous setup, each user is assigned samples from all three classes, in equal proportion. In the heterogeneous setup, each user is assigned samples from two out of three classes, with different number of samples per class and per user. The data distributions are visualized in Figure \ref{fig:data}. We set $\rho = 10$ in the homogeneous setup and $\rho = 100$ in the heterogeneous setup, due to different data distributions across users, to enforce consensus more strongly.

\begin{figure}[tp]
    \centering
    \begin{tabular}{cc}
        \includegraphics[width=0.45\linewidth]{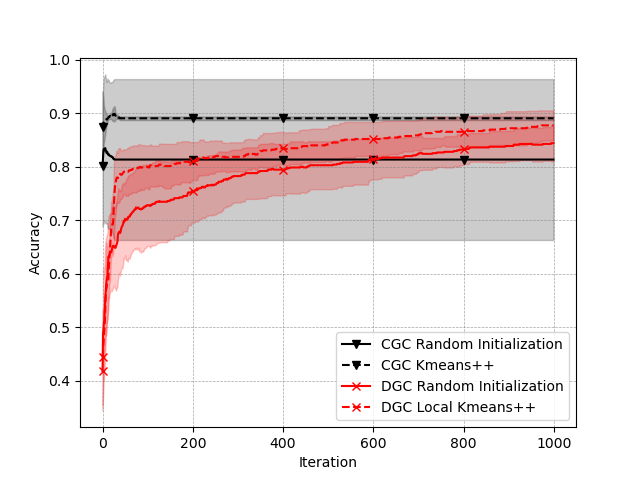}
        &
        \includegraphics[width=0.45\linewidth]{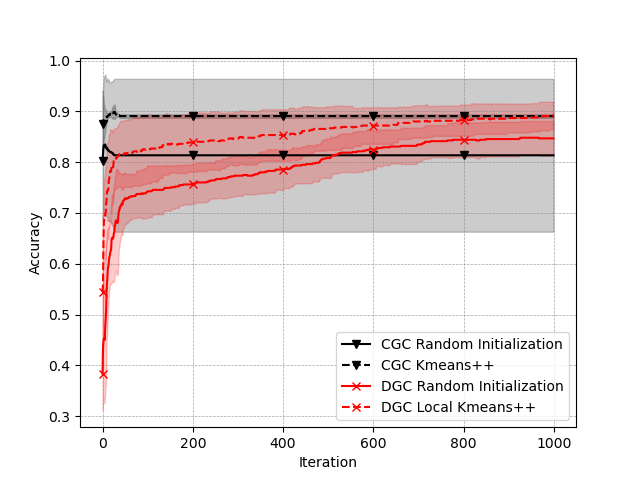}
    \end{tabular}
    \caption{Performance of methods on homogeneous and heterogeneous data.}
    \label{fig:acc-iter}
\end{figure}

To test the robustness to initialization of our method, we apply our DGC-KM method with two different center initializations: \emph{random} and \emph{local $K$-means++}. For random initialization, each user selects three centers uniformly at random from their local data. For local $K$-means++, each user initializes their centers using the $K$-means++ scheme on their local data. To compare the sensitivity of our algorithm, we use the CGC method from \cite{pmlr-v162-armacki22a}, also using the $K$-means cost. CGC is also initialized using random and $K$-means++ initialization, with the difference being that CGC chooses samples from the entire dataset, as it is a centralized algorithm. We run both methods for $T = 1000$ iterations, on both homogeneous and heterogeneous data.\footnote{Note that the distinction between homogeneous and heterogeneous data is irrelevant for CGC, as it is a centralized algorithm with access to all data.} We measure the performance via clustering \emph{accuracy}, i.e., by comparing the true labels to the ones produced by the clustering methods, accounting for label permutation. We repeat the experiments across $5$ runs and present the average accuracy. For our distributed method, we additionally average the accuracy across users. The results are presented in Figure \ref{fig:acc-iter}. The solid lines represent accuracy per iteration using random initialization, while dashed lines represent the performance using $K$-means++. We can see that our DGC-KM method shows less sensitivity to center initialization compared to CGC-KM, with the gap in performance of our method under different initialization much smaller than that of CGC-KM. This phenomena has previously been observed in \cite{dist-clust-wsn,armacki2024unified}, where it was noted that distributed clustering algorithms are less sensitive to initialization, compared to their centralized counterparts, as they in essence perform $m$ initialization, one per each user, whereas the centralized algorithms only perform one initialization. Combined with the effects of consensus, these multiple initializations help mitigate the effects of bad initialization at some users and lead the algorithm to a solution of better quality.  

Another interesting observation from Figure \ref{fig:acc-iter} is that the performance of DGC-KM is improved under local $K$-means++ initialization, even when the data across users is heterogeneous. This leads to a natural question of can the benefits of $K$-means++ initialization be further exploited in the distributed setup, if users are allowed to collaborate during the initialization phase? To answer this question, we design a novel distributed center initialization algorithm, dubbed Distributed K-Means+Clustering (DKM+C). The algorithm runs for a user-specified number of communication rounds $L \geq 0$, with $L = 0$ corresponding to each user initializing their centers by performing $K$-means++ on their local data. If $L \geq 1$, then in round $l = 1,\ldots,L$, users first exchange their centers from round $l - 1$ with their neighbours. Next, in order to account for potential label mismatch among different users, each user locally runs the $K$-means clustering algorithm on their own and their neighbours centers, i.e., on the dataset $\{x_i^{l-1}(k),\: x_j^{l-1}(k): \: k \in [K], \: j \in \calN_i \} \subset \R^d$, of size $(|\calN_i|+1)K$. The new centers $\bx_i^l$ are the centers returned by $K$-means. The steps are then repeated, until all $L$ communication rounds are performed. The proposed initialization scheme is summarized in Algorithm \ref{alg:dist-km++}. The idea behind the proposed scheme is to combine the power of $K$-means++ with local communications, to produce center initializations carrying more information than purely local initialization. Compared to some existing distributed $K$-means schemes, e.g., \cite{dist-fuzzy}, where users are required to run the max-consensus algorithm until convergence a total of $2k$ times, we provide a communication-efficient algorithm, that only communicates for a fixed number of rounds, while performing center inference locally, trading communication for computation.         

\begin{algorithm}[!tb]
\caption{DKM+C}
\label{alg:dist-km++}
\begin{algorithmic}[1]
   \REQUIRE{Number of centers $K$ and communication rounds $L \geq 0$.}
   \FOR{all users $i$ in parallel, in communication round $l = 0,1,\ldots,L$}
        \IF{l = 0}
            \STATE Produce $\bx_i^l$ by performing $K$-means++ on local data; 
        \ENDIF
        \STATE Exchange centers $\bx_i^{l-1},\bx_j^{l-1}$ with neighbours $j \in \calN_i$;
        \STATE Produce new centers $\bx_i^l$, by running $K$-means on $\{x_i^{l-1}(k), \: x_j^{l-1}(k): \: k \in [K], \: j \in \calN_i\}$;
   \ENDFOR
   \STATE Initialize centers via $\bx_i^0 \leftarrow \bx^L_i$, for all $i \in [m]$.
    \end{algorithmic}
\end{algorithm}

\begin{figure}[tp]
    \centering
    \begin{tabular}{cc}
        \includegraphics[width=0.49\linewidth]{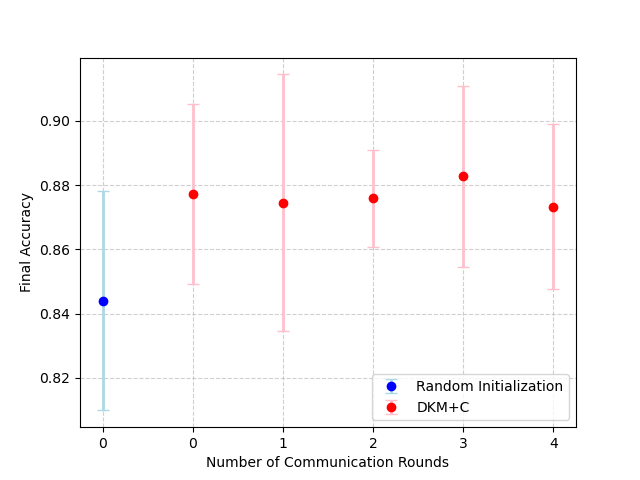}
        &
        \includegraphics[width=0.49\linewidth]{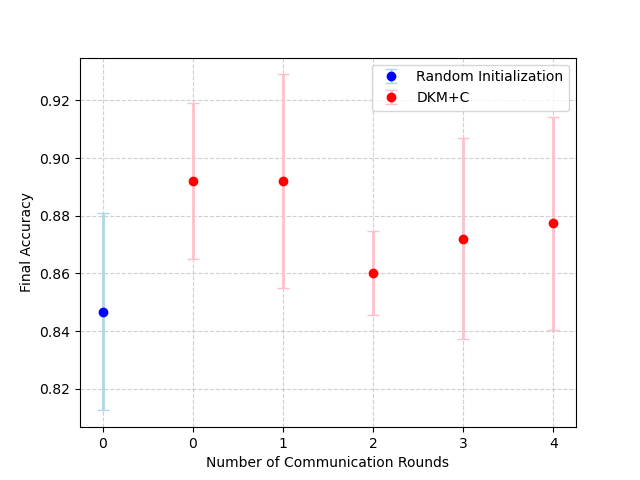}
    \end{tabular}
    \caption{Performance of methods on homogeneous and heterogeneous data.}
    \label{fig:acc-init}
\end{figure}

We test the impact of our new initialization scheme, by running DGC-KM for $T = 1000$ iterations, using random initialization and DKM+C with $L = \{0,1,2,3,4\}$, on both homogeneous and heterogeneous data. We then report the final accuracy of our method, averaged across $5$ runs. We again use $\rho = 10$ for homogeneous and $\rho = 100$ for heterogeneous data. The results are presented in Figure \ref{fig:acc-init}. The $x$ axis represents the number of communication rounds, while the $y$-axis represents the final accuracy obtained by DGC-KM. We can see that increasing the number of communication rounds benefits the initialization in both homogeneous ($L = 3$) and heterogeneous ($L = 1$) setup. More importantly, the initialization scheme outperforms the random initalization in both setups, showing clear improvements.   

\section{Conclusion}\label{sec:conclusion}

In this work we studied the effects of initialization on the performance of distributed clustering method originally proposed in \cite{armacki2024unified}. We demonstrated through experiments on real data that the performance of DGC-$\mathcal{F}_\rho$ is more robust to initialization compared to centralized gradient clustering method from \cite{pmlr-v162-armacki22a}. Next, we propose an initialization scheme, dubbed DKM+C, inspired by $K$-means++, which is shown to improve the performance of DGC-$\mathcal{F}_\rho$ compared to the baseline random initialization. Future work includes a rigorous theoretical analysis of the benefits of the proposed initialization scheme, as well as studying a version of the DGC-$\mathcal{F}_\rho$ method with a time-varying penalty $\rho_t$, such that $\rho_t \rightarrow \infty$, as $t \rightarrow \infty$.

\bibliographystyle{ieeetr}
\bibliography{bibliography}

\end{document}

%% file: macros.tex
\newcommand{\bx}{\mathbf{x}}

\newcommand{\R}{\mathbb{R}}
\newcommand{\D}{\mathcal{D}}
\newcommand{\C}{\mathcal{C}}
\newcommand{\N}{\mathbb{N}}
\newcommand{\calN}{\mathcal{N}}